\documentclass[preprint,review,12pt]{elsarticle}

\usepackage[utf8]{inputenc} 

\usepackage{amsmath, amsthm, amssymb}
\usepackage[bottom]{footmisc}
\usepackage{graphicx}
\usepackage{longtable}
\usepackage{pdfpages}
\usepackage{multirow}
\usepackage{subfigure}
\usepackage{algorithm}
\usepackage{algorithmic}
\usepackage{url}
\usepackage{epstopdf}
\def\cX{{\cal X}}

\def\cN{{\cal N}}

\def\cG{{\cal G}}

\def\bu{{\mbox{\boldmath $u$}}}
\def\bv{{\mbox{\boldmath $v$}}}
\def\bw{{\mbox{\boldmath $w$}}}
\def\bx{{\mbox{\boldmath $x$}}}
\def\bz{{\mbox{\boldmath $z$}}}

\def\bmu{{\mbox{\boldmath $\mu$}}}
\def\bLambda{{\mbox{\boldmath $\Lambda$}}}
\def\bSigma{{\mbox{\boldmath $\Sigma$}}}
\def\bPsi{{\mbox{\boldmath $\Psi$}}}
\def\bI{{\mbox{\boldmath $I$}}}

\def\btheta{{\mbox{\boldmath $\theta$}}}

\journal{Pattern Recognition}

\begin{document}

\begin{frontmatter}



\title{Adaptive Mixtures of Factor Analyzers}


\author[nkucmpe]{Heysem Kaya\corref{cor1}}
\ead{heysem@boun.edu.tr}
\author[bouncmpe]{Albert Ali Salah}
\ead{salah@boun.edu.tr}
\cortext[cor1]{Corresponding author}
\address[nkucmpe]{Department of Computer Engineering, Çorlu Faculty of Engineering \\ Namık Kemal University, 59860, \c{C}orlu, Tekirda\u{g}, TURKEY}
\address[bouncmpe]{Department of Computer Engineering \\ Bo\u{g}azi\c{c}i University, 34342, Bebek, \.{I}stanbul, TURKEY}

\begin{abstract}
A mixture of factor analyzers is a semi-parametric density estimator that generalizes the well-known mixtures of Gaussians model by allowing each Gaussian in the mixture to be represented in a different lower-dimensional manifold. This paper presents a robust and parsimonious model selection algorithm for training a mixture of factor analyzers, carrying out simultaneous clustering and locally linear, globally nonlinear dimensionality reduction. Permitting different number of factors per mixture component, the algorithm adapts the model complexity to the data complexity. We compare the proposed algorithm with related automatic model selection algorithms on a number of benchmarks. The results indicate the effectiveness of this fast and robust approach in clustering, manifold learning and class-conditional modeling.

\end{abstract}


\end{frontmatter}


\section{Introduction}
\label{sec:intro}
Mixture models have a widespread use in various domains of machine learning and signal processing for supervised, semi-supervised and unsupervised tasks~\cite{Moerland00dissertation,mclachlan2000finite}. However, the model selection problem remains to be one of the challenges and there is a need for efficient and parsimonious automatic model selection methods~\cite{Jain2010}.

Let $\bx$ denote a random variable in $\mathbb{R}^d$. A mixture model represents the distribution of $\bx$ as a mixture of $K$ component distributions:
\begin{equation}
p(\bx)= \sum_{k=1}^K{p\left(\bx|\cG_k\right)p\left(\cG_k\right)},
\label{eq:MixtureDistrib}
\end{equation}
where $\cG_k$ correspond to components, and $p\left(\cG_k\right)$ are the prior probabilities of the components. $p\left(\cG_k\right)$ are also called the \textit{mixture proportions}, and sum up to unity. The likelihood term, expressed by $p\left(\bx|\cG_k\right)$, can be modeled by any distribution. In this paper we focus on Gaussians:
\begin{equation}
p\left(\bx|\cG_k\right)\sim\cN(\bmu_k,\bSigma_k),
\label{eq:Gaussian}
\end{equation} 
where $\bmu_k$ and $\bSigma_k$ denote the mean and covariance of the $k^{th}$ component distribution, respectively. The number of parameters in the model is primarily determined by the dimensionality of the covariance matrix, which scales quadratically with the feature dimensionality $d$. When this number is large, overfitting becomes an issue. Indeed, one of the most important problems of model-based clustering methods is that they are over-parametrized in high-dimensional spaces~\cite{bouveyron2014model}. One way of keeping the number of parameters small is to constrain the covariance matrices to be tied (shared) across components, which assumes similar shaped distributions in the data space, and is typically unjustified. Another approach is to assume that each covariance matrix is diagonal or spherical, but this means valuable correlation information will be discarded. 

It is possible to keep a low number of parameters for the model without sacrificing correlation information by adopting a factor analysis approach. Factor Analysis (FA) is a latent variable model, which assumes the observed variables are linear projections of a small number of independent factors $\bz$ with additive Gaussian noise: 
\begin{equation}
\bx = \bLambda\bz + \bu, \bz\sim \cN(0, \bI), \bu\sim \cN(0, \bPsi),  
\label{eq:FAGen1}
\end{equation}
where $\bLambda$ is a $d \times p$ \textit{factor loading matrix} and $\bPsi$ is a diagonal uniquenesses matrix representing the common sensor noise. Subsequently, the covariance matrix in Eq.~\ref{eq:Gaussian} is expressed as $\bSigma_k = \bLambda_k\bLambda_k^T + \bPsi$, effectively reducing the number of parameters from $O(d^2)$ to $O(dp)$, with $p<<d$. If each Gaussian component is expressed in a latent space, the result is a mixture of factor analyzers (MoFA).

Given a set of data points, there exists Expectation-Maximization (EM) approaches to train MoFA models~\cite{mclachlan2000finite,Ghahramani97EM4MFA}, but these approaches require the specification of hyper-parameters like the number of clusters and the number of factors per component. For the model selection problem of MoFA, an incremental algorithm (IMoFA) was proposed in~\cite{Salah04imofa}, where factors and components were added to the mixture one by one. The model complexity was monitored on a separate validation set.

In this study, we propose a fast and parsimonious model selection algorithm called \textit{Adaptive Mixture of Factor Analyzers} (AMoFA). Similar to IMoFA, AMoFA is capable of adapting a mixture model to data by selecting an appropriate number of components and factors per component. However, the proposed AMoFA algorithm deals with two shortcomings of the IMoFA approach: 1) Instead of relying on a validation set, AMoFA uses a Minimum Message Length (MML) based criterion to control model complexity, subsequently using more training samples in practice. 2) AMoFA is capable of removing factors and components from the mixture when necessary. We test the proposed AMoFA approach on several benchmarks, comparing its performance with IMoFA, with a variational Bayesian MoFA approach~\cite{Ghahramani1999}, as well as with the popular Gaussian mixture model selection approach based on MML, introduced by Figueiredo and Jain~\cite{Figueiredo2002}. We show that the proposed approach is parsimonious and robust, and especially useful for high-dimensional problems.

The layout of the paper is organized as follows. In the next section we review related work in model selection.
AMoFA algorithm is introduced in Section~\ref{sec:amofa}. Experimental results are presented in Section~\ref{sec:experiments}. Section~\ref{lbl:conclusions} discusses our findings, and concludes with future directions.
\section{Related Work}
\label{lbl:relatedwork}
There are numerous studies for mixture model class selection. These include using information theoretical trade-offs between likelihood and model complexity~\cite{AkaikeAIC,SchwarzBIC,Rissanen1983,Rissanen:InCSM2007,Wallace87MML}, greedy approaches~\cite{verbeek2003efficient,Salah04imofa} and full Bayesian treatment of the problem~\cite{ Ghahramani1999,Rasmussen2000,GomesIncLearnNonpBayMMCVPR08,Shi2011}. A brief review of related automatic model selection methods is given in Table~\ref{tab:relatedwork}, a detailed treatment can be found in~\cite{bouveyron2014model}. Here we provide some detail on the most relevant automatic model selection methods that are closely related to our work.

\begin{table}[htbp]
  \centering
  \caption{Automatic Mixture Model Selection Approaches}
    \begin{scriptsize}
    \begin{tabular}{lll}
       \hline 
       {\footnotesize Work}  & {\footnotesize Model Selection} & {\footnotesize Approach} \\
       \hline
       Ghahramani \& Beal (1999)~\cite{Ghahramani1999} & Variational Bayes & Incremental  \\      
       Pelleg \& Moore (2000)~\cite{Pelleg2000XM} & MDL   & Incremental   \\
       Rasmussen (2000)~\cite{Rasmussen2000} & MC for DPMM & Both\\
       Figueiredo \& Jain (2002)~\cite{Figueiredo2002} & MML   & Decremental   \\
       Verbeek et al. (2003)~\cite{verbeek2003efficient} & Fixed iteration & Incremental \\
       Law et al. (2004)~\cite{LawFigJainPAMI04} & MML   & Decremental   \\
       Zivkovic \& v.d. Heijden (2004)~\cite{Zivkovic04RULFMM} & MML   & Decremental  \\
       Salah \& Alpaydin (2004)~\cite{Salah04imofa} & Cross Validation & Incremental \\
       Shi \& Xu (2006)~\cite{ShiMoFAICANN06} & Bayesian Yin-Yang & Both\\
       Constantinopoulos et al. (2007)~\cite{ConstantinopoulosVarComSplitNN07} & Variational Bayes & Incremental \\
       Gomes et al. (2008)~\cite{GomesIncLearnNonpBayMMCVPR08} & Variational DP & Incremental \\
       Boutemedjet et al. (2009)~\cite{Boutemedjet2009} & MML   & Decremental  \\
       Gorur \& Rasmussen (2009)~\cite{Gorur09SIUNonParMFA} & MC for DPMM & Both \\
       Shi et al. (2011)~\cite{Shi2011} & Bayesian Yin-Yang & Both\\
       Yang et al. (2012)~\cite{yang2012robust} & Entropy Min. & Decremental  \\
       Iwata et al. (2012)~\cite{Iwata1206.1846} & MC for DPMM & Both\\
       Fan \& Bouguila (2013)~\cite{FanVBCSNeuCom2013} & Variational DP & Both \\
       Fan \& Bouguila (2014)~\cite{FanOVBDMFSNeuCom2014} & Variational Bayes & Incremental \\
       Kersten (2014)~\cite{Kersten2014} & MML   & Decremental  \\
       \hline
       
       \end{tabular}%
       \end{scriptsize}
  \label{tab:relatedwork}%
\end{table}%

In one of the most popular model selection approaches for Gaussian mixture models (GMMs), Figueiredo and Jain proposed to use an MML criterion for determining the number of components in the mixture, and shown that their approach is equivalent to assuming Dirichlet priors for mixture proportions~\cite{Figueiredo2002}. In their method, a large number of components (typically 25-30) is fit to the training set, and these components are eliminated one by one. At each iteration, the EM algorithm is used to find a converged set of model parameters. The algorithm generates and stores all intermediate models, and selects one that optimizes the MML criterion. 

The primary drawback of this approach is the curse of dimensionality. For a $d$-dimensional problem, fitting a single full-covariance Gaussian requires $O(d^2)$ parameters, which typically forces the algorithm to restrict its models to diagonal covariances in practice. We demonstrate empirically that this approach (unsupervised learning of finite mixture models - ULFMM) does not perform well in practice for problems with high dimensionality, regardless of its abundant use in the literature.

\begin{figure}
\begin{center} 
\includegraphics[width=0.95\columnwidth,bb=0 0 769 427]{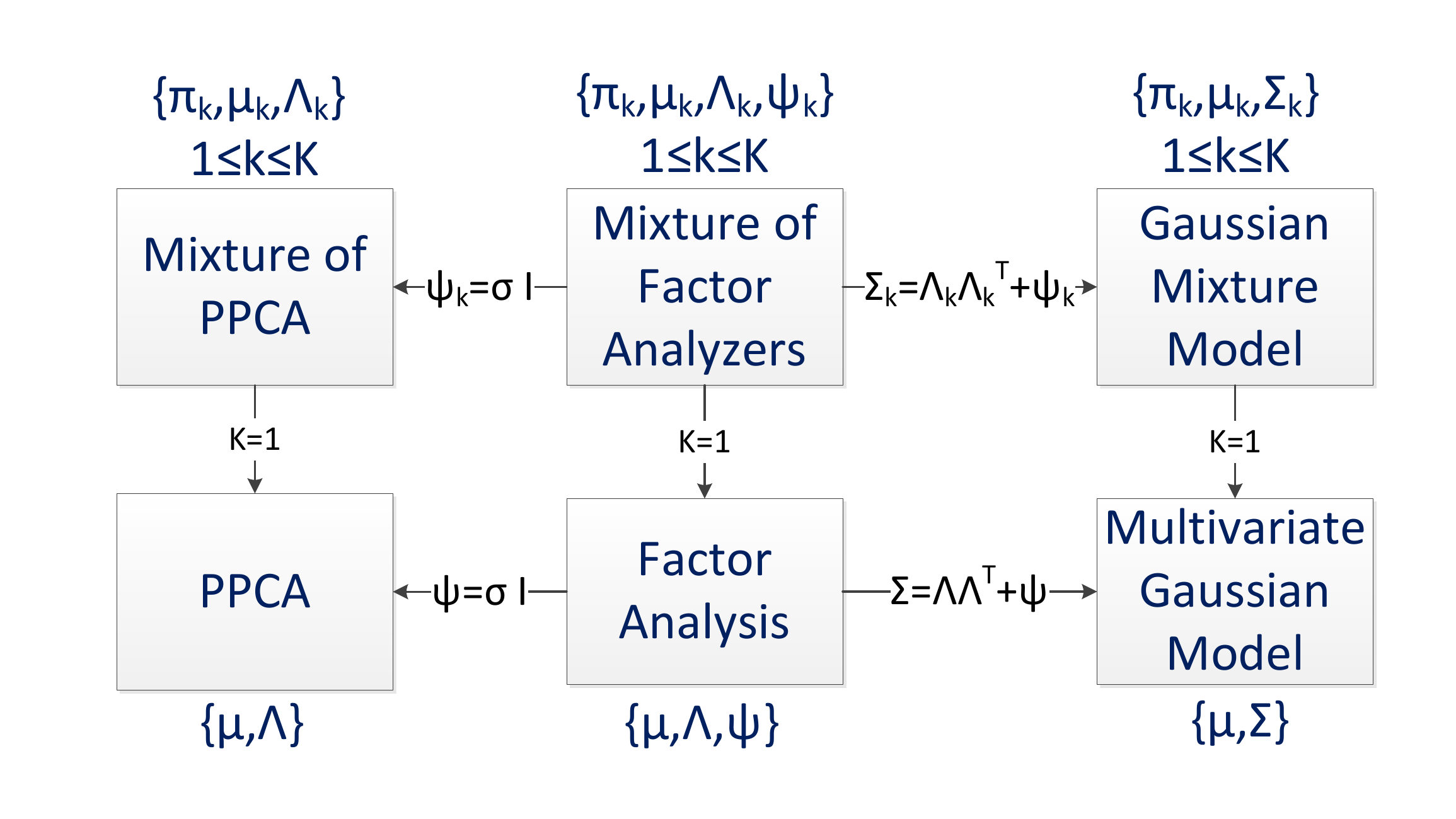}
\caption{Relationship of MoFA with some well known latent variable and mixture models. Model parameters are given in curly brackets. $\pi$: ($1\times K$) component priors, $\mu$: (1$\times d$) component mean, $\lambda$: ($p\times d$) factor loading matrix, $\Psi$: ($1\times d$) diagonal noise variances (uniqueness),  $\Sigma$: ($d\times d$) component covariance.) $K$ denotes the number of components, $d$ the feature dimensionality, and $p$ the subspace dimensionality with $p<<d$.}
\label{fig:mofa_relations}
\end{center}
\end{figure}
Using the parsimonious factor analysis representation described in Section~\ref{sec:intro}, it is possible to explore many models that are between full-covariance and diagonal Gaussian mixtures in their number of parameters. The resulting mixture of factor analysers (MoFA) can be considered as a noise-robust version of the mixtures of probabilistic principal component analysers (PPCA) approach~\cite{Tipping99MPPCA}. Figure~\ref{fig:mofa_relations} summarizes the relations between the mixture representations in this area. 

If we assume that the latent variables of each component $\cG_k$ in a MoFA model is distributed unit normal (${\cal N}(0,I)$) in the latent space, the corresponding data in the feature space are also Gaussian distributed: 
\begin{equation}
p\left( {\cal X}|\bz,\cG_k \right)={\cal N} \left(\bmu_k + \bLambda_k \bz,\bPsi_k \right),
\label{eq:MFAlik}
\end{equation}
where $\bz$ denotes the latent factor values. The mixture distribution of $K$ factor analyzers is then given as~\cite{Ghahramani97EM4MFA}:
\begin{equation}
\label{eq:MFAdensity}
p({\cal X})= \sum_{k=1}^K \int p({\cal X}|\bz,\cG_k)p(\bz|\cG_k)p(\cG_k) dz.
\end{equation}
The EM algorithm is used to find maximum likelihood solutions to latent variable models~\cite{Dempster77MLEM}, and it can be used for training a MoFA~\cite{Ghahramani97EM4MFA}. Since EM does not address the model selection problem, it requires the number of components and factors per component to be fixed beforehand. 

Ghahramani and Beal~\cite{Ghahramani1999} have proposed a variational Bayes scheme (VBMoFA) for model selection in MoFA, which allows the local dimensionality of components and their total number to be automatically determined. In this study, we use VBMoFA as one of the benchmarks. 

To alleviate the computational complexity of the variational approach, a greedy model selection algorithm was proposed by Salah and Alpaydın~\cite{Salah04imofa}. This incremental approach (IMoFA) starts by fitting a single component - single factor model to the data and adds factors and components in each iteration using fast heuristic measures until a convergence criterion is met. The algorithm allows components to have as many factors as necessary, and uses a validation set to stop model adaptation, as well as to avoid over-fitting. This is the third algorithm we use to compare with the proposed approach, which we describe in detail next.

\section{Adaptive Mixtures of Factor Analyzers}
\label{sec:amofa}
We briefly summarize the proposed adaptive mixtures of factor analyzers (AMoFA) algorithm first, and then describe its details. Given a dataset $\cX$ with $N$ data points in $d$ dimensions, the AMoFA algorithm is initialized by fitting a 1-component, 1-factor mixture model. Here, the factor is initialized from the leading eigenvector of the covariance matrix i.\,e. the principal component of the data. At each subsequent step, the algorithm considers adding more components and factors to the mixture, running EM iterations to find a parametrization. During the M-step of EM, an MML criterion is used to determine whether any weak components should be annihilated. Apart from this early component annihilation, the algorithm incorporates a second decremental scheme. When the incremental part of the algorithm no longer improves the MML criterion, a downsizing component annihilation process is initiated and all components are eliminated one by one. Similar to ULFMM, each intermediate model is stored, and the algorithm outputs the one giving the minimum message length. Figure~\ref{fig:amofa} summarizes the proposed algorithm. 

\begin{figure}[htbp]
\caption{\label{fig:amofa}Outline of the AMoFA algorithm}
\begin{center}
\fbox{
\begin{minipage}{8cm}
\begin{tabbing}
123\=456\=789\=123\=\kill
algorithm {\bf AMoFA}(training set $\cX$)\\
\> /*Initialization*/\\
\> [$\bLambda, \bmu, \Psi$] $\gets$ train a 1-component, 1-factor model\\
\> {\bf repeat} \\
\> /*Perform a single split*/\\
\> \> x $\gets$ Select a component for splitting via Eq.~(\ref{eq:mardiamvkurt})\\
\> \> [$\bLambda_{1}, \bmu_{1}, \Psi_{1}, \pi_{1}$] $\gets$ MML\_EM(split x). \\
\> \> actionML($1$) $\gets$ ML($\bLambda_{1}, \bmu_{1}, \Psi_{1}, \pi_{1}$) via  Eq.~(\ref{eq:FJMMLAdoptedAmoFAFinal})\\
\> /*Perform a single factor addition*/\\
\> \> y $\gets$ Select a component to add a factor\\
\> \> [$\bLambda_{2}, \bmu_{2}, \Psi_{2}, \pi_{2}$] $\gets$ MML\_EM(add factor to y). 
\\
\> \> actionML($2$) $\gets$ ML($\bLambda_{2}, \bmu_{2}, \Psi_{2}, \pi_{2}$) via  Eq.~(\ref{eq:FJMMLAdoptedAmoFAFinal})\\
 \> /*Select the best action*/\\
\> \> z $\gets$ arg min(actionML(1),actionML(2))\\ 
 \> /*Update the parameters*/\\
\> \> [$\bLambda, \bmu, \Psi, \pi$] $\gets$ [$\bLambda_{z}, \bmu_{z}, \Psi_{z}, 
\pi_{z}$]\\
\> {\bf until} MML decrease $< \epsilon$\\
\> /*Annihilation starts with $k=K$ components*/\\
\> {\bf while} $k > 1$\\
\> /*Select the weakest component for annihilation*/\\
\> \> [$\bLambda_{k}, \bmu_{k}, \Psi_{k}, \pi_{k}$] $\gets$ EM(annihilate component). \\
\> \> k = k - 1\\
\> {\bf end} \\
\> /*Select $l$ that minimizes MML criterion in  Eq.~(\ref{eq:FJMMLAdoptedAmoFAFinal})*/\\
\> {\bf return} [$\bLambda_{l}, \bmu_{l}, \Psi_{l}, \pi_{l}$]\\
{\bf end}
\end{tabbing}
\end{minipage}
}
\end{center}
\end{figure}

\subsection{The Generalized Message Length Criterion}
To allow local factor analyzers to have independent latent dimensionality, the MML criterion given in Figueiredo and Jain~\cite{Figueiredo2002} should be generalized accordingly to reflect the individual code length of components:
\begin{equation}
\label{eq:FJMMLAdoptedAmoFA}
\begin{split}
{\cal L}(\btheta,{\cal X}) =&  \sum_{k:\pi_k>0} \frac{C_k}{2} \log (\frac{N\pi_k}{12}) + \frac{K_{nz}}{2} \log \frac{N}{12} +
\\ & \sum_{k:\pi_k>0}\frac{(C_k+1)}{2}-\log p({\cal X}|\btheta),
\end{split}
\end{equation}
where $C_k$ denotes the number of parameters for component $k$, $\cal X$ represents the dataset with $N$ data items, $\btheta$ the model parameters, and $K_{nz}$ represents the number of non-zero weight components. The first three terms in Eq.~\ref{eq:FJMMLAdoptedAmoFA} comprise the code length for real valued model parameters, the fourth term is the model log-likelihood. We propose to include the code length for integer hyper parameters, namely $K_{nz}$ and component-wise latent dimensionalities $\{p_k\}$, such that the encoding becomes decodable as required by MDL theory~\cite{Rissanen1983,Rissanen:InCSM2007}. 
For this purpose, we use  Rissanen's \textit{universal prior for integers}~\cite{Rissanen:InCSM2007}:
\begin{equation}
\label{eq:universalprior}
w^*(k)=c^{-1}2^{-log^*k},
\end{equation}
which gives the (ideal) code length
\begin{equation}
\label{eq:idealcodelength}
L^*(k)=log\ 1/w^*(k)=log^*(k) + log\ c,
\end{equation}
where $log^*(k)= logk + loglogk + ... $ is n-fold logarithmic sum with positive terms, $c$ is the normalizing sum $\sum_{k>0}2^{-log^*k}$ that is tightly approximated as\ $c=2.865064$~\cite{Rissanen:InCSM2007}.
$log^*(k)$ term in Eq.~\ref{eq:idealcodelength} can be computed via a recursive algorithm. We finally obtain $L^*(K_{nz})$, the cost to encode the number of components, and similarly $\sum_{k:\pi_k>0} L^*(p_k)$, the cost to encode the local dimensionalities $\{p_k\}$ and add them to eq.\ (\ref{eq:FJMMLAdoptedAmoFA}) to obtain a message length criterion:
\begin{equation}
\label{eq:FJMMLAdoptedAmoFAFinal}
\begin{split}
{\cal L}(\btheta,{\cal X}) =&  \sum_{k:\pi_k>0} \frac{C_k}{2} \log (\frac{N\pi_k}{12}) + \frac{K_{nz}}{2} \log \frac{N}{12} 
\\ & + \sum_{k:\pi_k>0}\frac{(C_k+1)}{2} -\log p({\cal X}|\btheta) 
\\ & +  L^*(K_{nz}) + \sum_{k:\pi_k>0} L^*(p_k)
\end{split}
\end{equation} 

\subsection{Component Splitting and Factor Addition}
Adding a new component by splitting an existing one involves two decisions: 
which component to split, and how to split it. AMoFA splits the 
component that looks least likely to a Gaussian, by looking at a multivariate kurtosis 
metric~\cite{MardiaMA}. For a multinormal distribution, the multivariate kurtosis 
takes the value $\beta_{2,d} = d(d+2)$, and if the underlying population is multivariate normal with mean 
$\bmu$, the sample counterpart of $\beta_{2,d}$, namely $b_{2,d}$, has an asymptotic distribution as the number of samples $N $ goes to infinity. Salah and Alpayd{\i }n~\cite{Salah04imofa} adapted this metric to the mixture model by using a ``soft count" $h^t_j\equiv E[\cG_j|\bx^t]$:
\begin{equation}
\label{eq:mardiamvkurt}
\gamma_j = \{b_{2,d}^j - d(d+2)\}{\left[ 
\frac{8d(d+2)}{\sum_{t=1}^{N}h^t_j}\right]^{-\frac{1}{2}} } 
\end{equation}
\begin{equation}
b_{2,d}^j = \frac{1}{\sum_{l=1}^{N}h^l_j}
        \sum_{t=1}^{N}h^t_j\left[(\bx^t-\bmu_j)^T \bSigma_j^{-1} 
(\bx^t-\bmu_j)\right]^{2}
\end{equation} 

The component with greatest $\gamma_j$ is selected for splitting. AMoFA runs a local, 2-component MoFA on the data points that fall under the component. To initialize the means of new components prior to MoFA fitting, we use the weighted sum of all eigenvectors of the local covariance matrix: $\bw=\sum_i^d \bv_i \lambda_i$, and set $\bmu_{new} = \bmu \pm \bw$, where $\bmu$ is the mean vector of the component to split.

The component having the largest difference between modeled and sample covariance is selected for factor addition. As in IMoFA, AMoFA uses the residual factor addition scheme. Given a component $\cG_j$ and a set of data points $\bx^t$ under it, the re-estimated points after projection to the latent subspace can be written as: $\tilde{\bx}^t_j = \bLambda_j E[\bz^t|\bx^t,\cG_j]$. The re-estimation error decreases with the number of factors used in $\bLambda_j$. The newly added column in the factor loading matrix, $\bLambda_{j,p+1}$, is selected to be the principal direction (the eigenvector with the largest eigenvalue) of the residual vectors $\tilde{\bx}^t_j - \bx^t_j$. This new factor is used in bootstrapping the EM procedure.
\subsection{Component Annihilation}
In a Bayesian view, the message length criterion (eq.\ (\ref{eq:FJMMLAdoptedAmoFAFinal})) adopted from Figueiredo and Jain~\cite{Figueiredo2002} corresponds to assuming a flat prior on component parameters $\btheta_k$, and a Dirichlet prior on mixture proportions $\pi_k$:
\begin{equation}
\label{eq:DirPriorPi}
p(\pi_1,\cdots,\pi_K) \propto exp\{\sum_{k=1}^{K_{nz}}-\frac{C_k}{2}log\pi_k \} = \prod_{k=1}^{K_{nz}}\pi_k^{-C_k/2}. 
\end{equation}
Thus, in order to minimize the adopted cost in eq.\ (\ref{eq:FJMMLAdoptedAmoFAFinal}), the M-step of EM is changed for $\pi_k$ : 
\begin{equation}
\label{eq:FJpiUpd}
\hat{\pi}_k^{new}=\frac{max\{0,(\sum_{i=1}^{N}h_{ik})-\frac{C_k}{2} \}}{\sum_{j=1}^{K_{nz}} max\{0,(\sum_{i=1}^{N}h_{ij})-\frac{C_k}{2}\}},
\end{equation}
which means that all components having a soft count ($N_k$) smaller than half the number of local parameters $C_k$ will be annihilated. This threshold enables the algorithm to get rid of components that do not justify their existence. In the special case of AMoFA, the number of parameters per component are defined as:
\begin{equation}
\label{eq:compannihthresh}
C_k = d*(p_k+2)+L^*(p_k),
\end{equation}
where $d$ is the original dataset dimensionality, $p_k$ is the local latent dimensionality of component $k$, and $L^*(p_k)$ is the code length for $p_k$. The additive constant $2$ inside the bracket accounts for the parameter cost of mean $\bmu_k$ and local diagonal uniquenesses matrix $\bPsi_k$. Finally, the localized annihilation condition to check at the M step of EM is simply $N_k < T_k = C_k/2$.

In AMoFA, we use an outer loop to drive the model class adaptation and an inner EM loop to fit a mixture of factor analyzer model with initialized parameters. The inner EM algorithm is an improved and more generalized version of ULFMM~\cite{Figueiredo2002}, where after parallel EM updates we select the weakest component and check $N_k < T_k$ for annihilation, as opposed to sequential component update approach (using Component-wise EM -$CEM^2$~\cite{celeux2001component}). Any time during EM, automatic component annihilation may take place. When the incremental progress is saturated, the downsizing component annihilation is initiated. The MML based EM algorithm and relevant details are given in~\ref{appsec:app_em_mofa}. 


\section{Experiments}
\label{sec:experiments}
\subsection{Evaluation Protocol for Clustering Performance}
We compare AMoFA with two benchmark algorithms on clustering, namely ULFMM algorithm from~\cite{Figueiredo2002}\footnote{The code is available at \url{http://www.lx.it.pt/~mtf}} and the IMoFA-L from~\cite{Salah04imofa}.

We use the Normalized Information Distance (NID) metric for evaluating clustering accuracy, as it possesses several important properties; in addition to being a metric, it admits an analytical adjustment for chance, and allows normalization to [0-1] range~\cite{Vinh:2010:ITM}. NID is formulated as:
\begin{equation}
\label{eq:NID}
1-\frac{MI(\bu,\bv)}{\max \{H(\bu),H(\bv)\}},
\end{equation}
where entropy $H(\bu)$ and the mutual information $MI(\bu,\bv)$ for clustering are defined as follows:
\begin{eqnarray}
\label{eq:clustering_entropy}
H(\bu) & = & -\sum_{i=1}^R \frac{a_i}{N} \log \frac{a_i}{N}, 
\\
\label{eq:clustering_mutualinfo}
MI(\bu,\bv) & =& \sum_{i=1}^R \sum_{j=1}^C \frac{n_{ij}}{N} \log \frac{n_{ij}/N}{a_i b_j/N^2},
\end{eqnarray}
Here, $a_i$ is the number of samples in cluster $i$, $n_{ij}$ is the number of samples falling into cluster $i$ in clustering $\bu$ and cluster $j$ in clustering $\bv$. MI is a nonlinear measure of dependence between two random variables. It quantifies how much information in bits the two variables share. 
We compute NID between the ground truth and the clusterings obtained by the automatic model selection techniques in order to give a more precise measure of clustering than just the number of clusters. When there is no overlap, NID is expected to be close to 0; higher overlap of clusters might result in higher average NID, though a relative performance comparison can still be achieved.

\subsection{Experiments on Benchmark Datasets for Clustering}
\label{subsec:clustexperiments}
We tested three algorithms, namely IMoFA-L, AMoFA and ULFMM on benchmark synthetic/real datasets for clustering. For maximum comparability with previous work, we used some synthetic dataset examples from Figueiredo and Jain~\cite{Figueiredo2002}, as well as from a Yang et al.'s study on automatic mixture model selection~\cite{yang2012robust}. 

AMoFA, as opposed to IMoFA and ULFMM, does not rely on random initialization. In IMoFA, the first factor is randomly initialized, and in ULFMM the initial cluster centers are assigned to randomly selected instances. AMoFA initializes the first factor from the principal component of the dataset. Similar to residual factor addition, this scheme can be shown to converge faster than random initializations. Given a dataset, a single simulation is sufficient to assess performance. 

\begin{figure*}
\centering
\includegraphics[width=1.15\linewidth,bb=0 0 659 392]{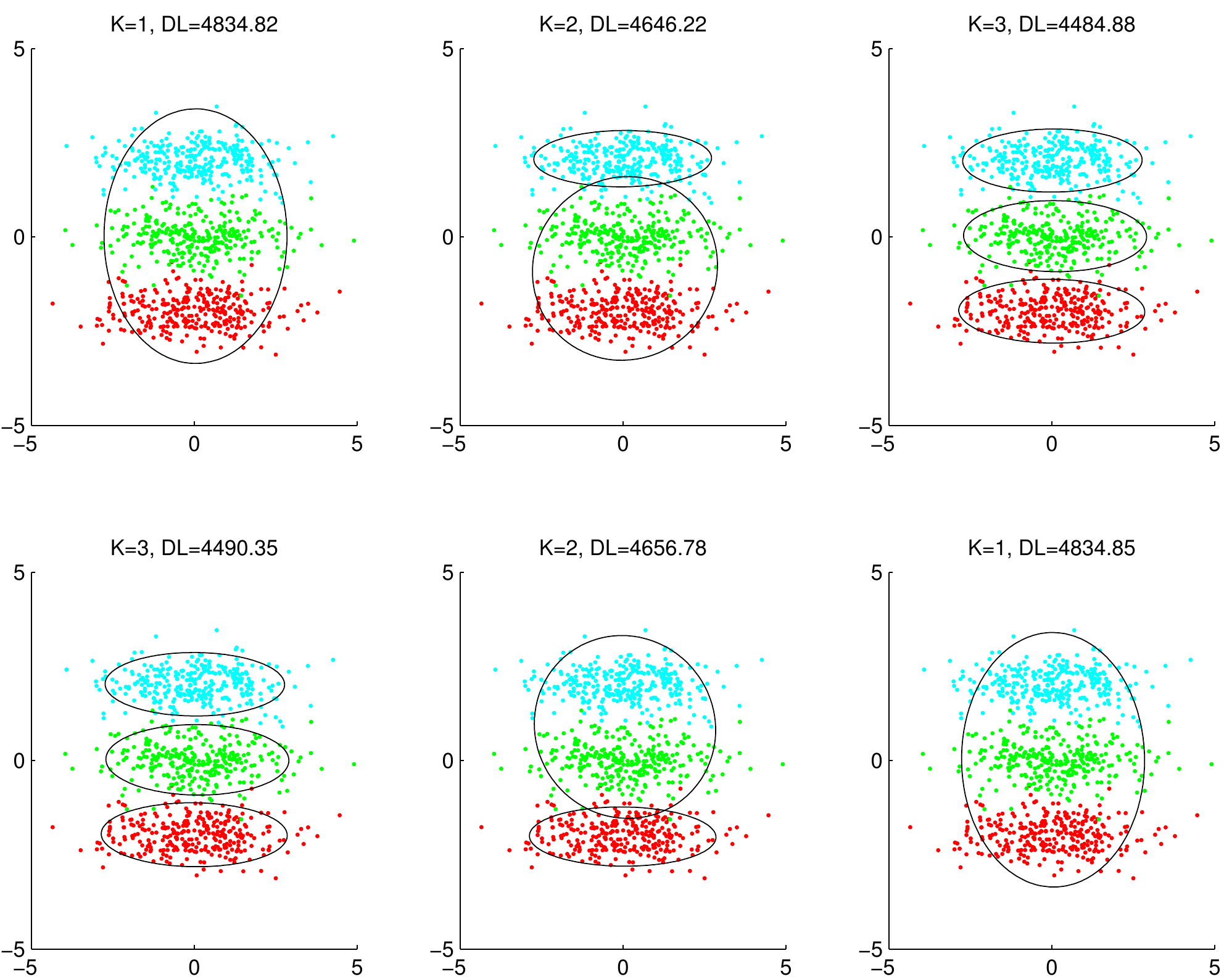}
\caption[3 Gaussians Progress]{\label{fig:3GaussiansProgress8}The evolution of AMoFA on a toy synthetic data. To keep the figure uncluttered, only the mixture models obtained at the end of adaptive steps are given. The initial step 
fits a single component-single factor model.
The first two iterations add components to the mixture, and the next one add a factor. 
The incremental phase stops when no (considerable) improvement in the message length is observed. Then, the algorithm starts to annihilate the components, until a single component is left. As expected, the DL in the decremental phase is higher, since components have two factors. Finally, the algorithm selects the 3-component solution having the minimum DL.}
\end{figure*}

Because of this deterministic property of AMoFA, we report the results with multiple datasets sampled from the underlying distribution, instead of sampling once and simulating multiple times. Unless stated otherwise, in the following experiments with synthetic datasets, 100 samples are drawn and the average results are reported. For ULFMM, we give initial number of clusters $K^{max}=20$ in all our simulations for clustering and use free full covariances. Moreover, the EM convergence threshold $\epsilon$ is set to $10^{-5}$ in all three methods. 
\paragraph{Example 1: 3 Separable Gaussians} To illustrate the evolution of the solution with AMoFA, we generated a mixture of three Gaussians having the same mixture proportions $\pi_1=\pi_2=\pi_3=1/3$ and the same covariance matrix $diag\{2,0.2\}$ with separate means $\bmu_1= [0,-2]' ,\bmu_2= [0,0]',\bmu_3= [0,2]'$. We generate 100 samples from the underlying distribution of 900 data points. This synthetic example is used in~\cite{Figueiredo2002, yang2012robust}.
Figure~\ref{fig:3GaussiansProgress8} shows the evolution of adaptive steps of AMoFA with found clusters shown in 2-std contour plots, and the description length (DL) is given above each plot. 

\paragraph{Example 2: Overlapping Gaussians} To test the approach for finding the correct number of clusters, we use a synthetic example very similar to the one used in~\cite{Figueiredo2002,yang2012robust}. Here, three of the four Gaussians overlap with the following generative model:
\\
$ \pi_1=\pi_2=\pi_3=0.3,\ \pi_4=0.1,\\  
\bmu_1=\bmu_2=[-4\ -4]',\ \bmu_3=[2\ 2]',\ \bmu_4=[-1\ -6]',\\ 
\bSigma_1=\begin{bmatrix}
.8&.5
\\
.5&.8
\end{bmatrix},\ 
\bSigma_2=\begin{bmatrix}
5&-2
\\
-2&5
\end{bmatrix},\ 
\bSigma_3=\begin{bmatrix}
2&-1
\\
-1&2
\end{bmatrix},\
\bSigma_4=\begin{bmatrix}
0.125&0
\\
0&0.125
\end{bmatrix}. $

\begin{figure*}
\centering
\begin{subfigure}
\centering
\includegraphics[width=0.49\linewidth,bb=0 0 560 420]{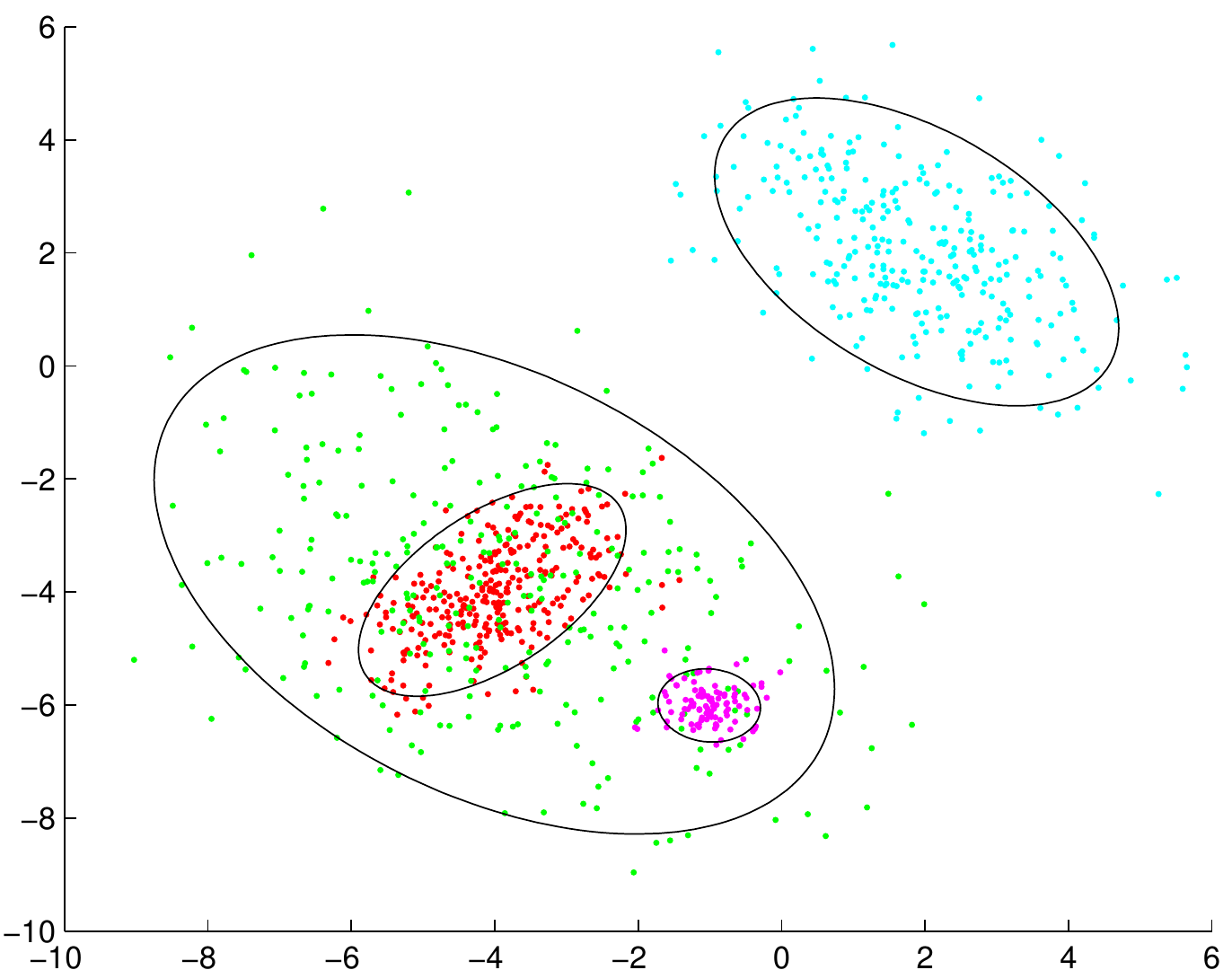}
\end{subfigure}%
\begin{subfigure}
\centering
\includegraphics[width=0.49\linewidth,bb=0 0 560 458]{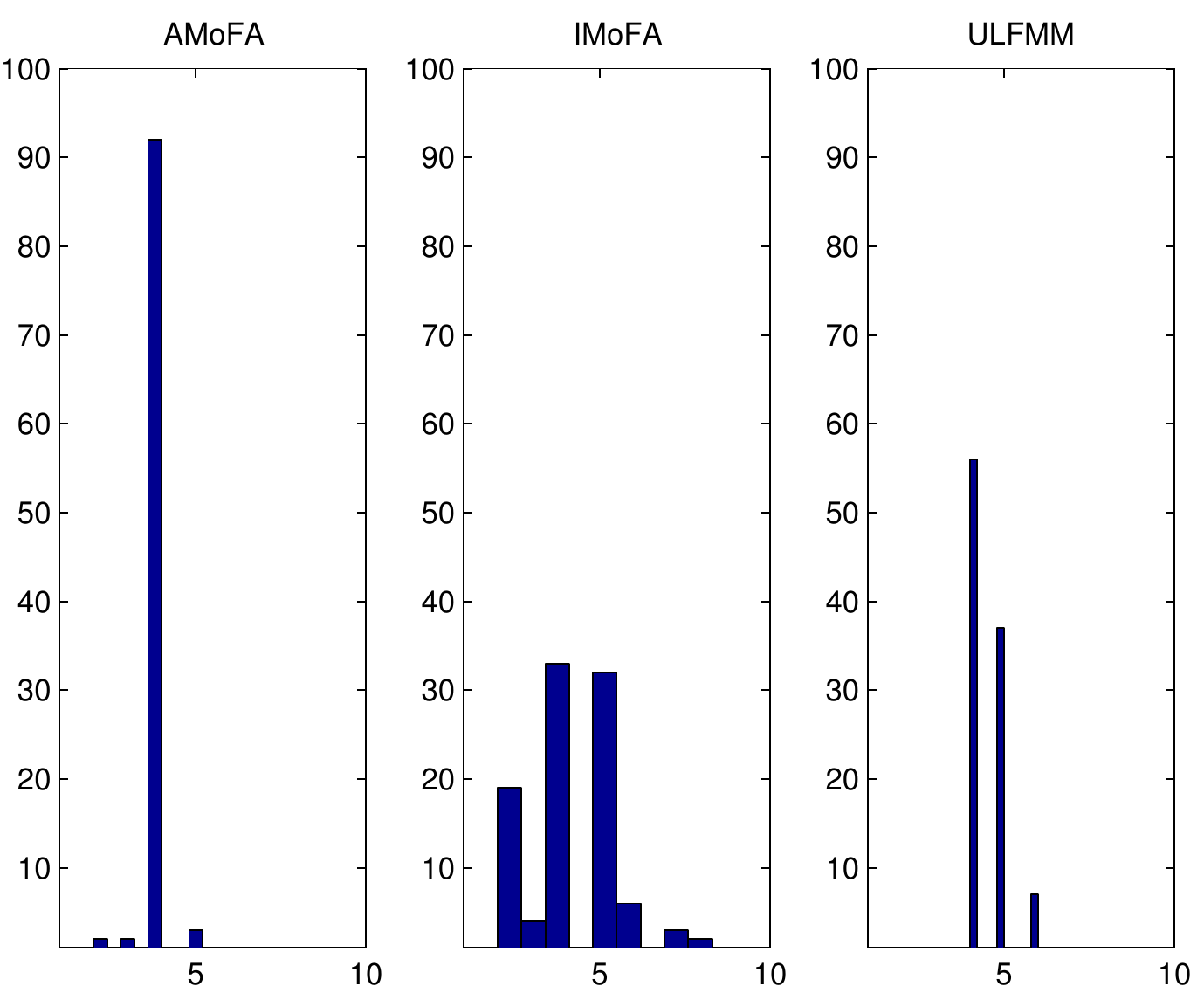}
\end{subfigure}
\caption{\label{fig:4OverlapGauss_result_hist}Overlapping Gaussians data. Left: A sample AMoFA result. The real labels are shown with colors and resulting AMoFA mixture model is shown with 2-std contour plot. Right: Histograms of number of clusters found by  AMoFA, IMoFA and ULFMM respectively.}
\end{figure*}

We use $N=1000$ data points. As in the previous example, we generate 100 random datasets. In figure~\ref{fig:4OverlapGauss_result_hist} left plot, the data are illustrated with a sample result of AMoFA. Out of 100 simulations, the accuracy of finding K*=4 is 92, 56, and 33 for AMoFA, ULFMM, and IMoFA, respectively. The histogram in figure~\ref{fig:4OverlapGauss_result_hist} right plot shows the distribution of number of automatically found clusters for three methods. Average NID over 100 datasets is found to be 0.2549, 0.2951, and 0.3377 for AMoFA, ULFMM and IMoFA, respectively. A  paired t-test (two tailed) on NID scores indicates that AMoFA performs significantly better than ULFMM with $p<10^{-5}$. 

\subsection[Modeling Class Conditional Densities]{Application to Classification: Modeling Class Conditional Densities}
\label{subsec:classifexperiments}
We compare AMoFA with three benchmark model selection algorithms, namely, VBMoFA
algorithm from~\cite{Ghahramani1999}, ULFMM
algorithm from~\cite{Figueiredo2002} and the IMoFA algorithm from~\cite{Salah04imofa}. As baseline, we use Mixture of Gaussians, where the data of each class are modeled by a single Gaussian with full (MoG-F) or diagonal (MoG-D) covariances. We compare the performances of the methods on classification tasks (via class-conditional modeling)
on nine benchmark datasets: The ORL face database with binary gender classification task~\cite{samaria1994parameterisation}, 16-class phoneme database from LVQ package of Helsinki University of Technology~\cite{LVQPAK}, the VISTEX texture database~\cite{Salah04imofa},
a 6-class Japanese Phoneme database\footnote{Pre-processed versions of VISTEX and Japanese Phoneme datasets that are used in this study can be accessed from \url{http://www.cmpe.boun.edu.tr/~kaya/jpn_vis.zip}}
~\cite{GurgenAUA94}, the MNIST dataset~\cite{lecun1998gradient}, and four datasets (Letter, Pendigits, Opdigits, and Waveform) from UCI ML Repository~\cite{UCIML}.
Table~\ref{tab:MoMdbs} gives some basic statistics about the databases. Except MNIST that has an explicit train and testing protocol, all experiments were carried out with 10-fold cross-validation. Simulations are replicated 10 times in MNIST, where we crop the 4 pixel padding around the images and scale them to $10\times 10$ pixels to obtain 100-dimensional feature vectors.

\begin{table}[htbp]
  \centering
  \caption[Datasets Used for Class Conditional Mixture Modeling]{Datasets Used for Class Conditional Mixture Modeling}
    \begin{tabular}{|c|c|c|c|}
    \hline
    Dataset & Dimensions & Classes & \# of Samples \\
    \hline
    ORL   & 256   & 2     & 400   \\
    LVQ   & 20    & 16    & 3858  \\
    OPT   & 60    & 10    & 4677  \\
    PEN   & 16    & 10    & 8992  \\
    VIS   & 169   & 10    & 3610  \\
    WAV  & 21    & 3     & 500  \\
    JPN   & 112   & 6     & 1200  \\
    LET   & 16 & 26 & 20000 \\
    MNT 	& 100 & 10 & 70000 \\
    \hline
    \end{tabular}%
  \label{tab:MoMdbs}%
\end{table}%

In the experiments, we trained separate mixture models for the samples of each class, and used maximum likelihood classification. We did not use informative class priors, as it would positively bias the results, and hide the impact of likelihood modeling. In Table~\ref{tab:allresults}, we provide accuracy computed over 10 folds, where all six approaches used the same protocol. 
ULFMM column reports performance of ULFMM models with free diagonal covariances, as full covariance models invariably give poorer results. 
\begin{table}[hbp]
 \centering
 \caption[Classification Accuracies for Class-Conditional Models]{Classification Performances for Class-Conditional Models. Significantly better results compared to the first runner up are shown in \textbf{bold}, where  * signifies 0.05 significance level, while ** corresponds to 0.01 significance level. If there are multiple best performers without pair-wise significant difference, they are shown in bold altogether.}
   \begin{tabular}{|c|c|c|c|}
   \hline
         & IMoFA-L~\cite{Salah04imofa} &  VBMoFA~\cite{Ghahramani1999} & AMoFA \\
  
   \hline
      {ORL}    & \textbf{97.8 $\pm$ 1.5} & 93.0 $\pm$ 2.8 & \textbf{97.5 $\pm$  1.2} \\
      {LVQ}    & \textbf{91.2 $\pm$ 1.9} & \textbf{91.3 $\pm$ 1.9} & 89.3 $\pm$ 1.6    \\
       {OPT}    & 91.1 $\pm$ 2.7  & \textbf{95.2 $\pm$ 1.8} & 93.8 $\pm$ 2.4\\
        {PEN}    & \textbf{97.9 $\pm$ 0.7} & \textbf{97.8 $\pm$ 0.6}  & \textbf{98.1 $\pm$ 0.6}\\
      {VIS}    & 69.3 $\pm$  4.6  & 67.1 $\pm$ 5.9 & \textbf{77.2 $\pm$ 4.6**} \\
      {WAV}    & 80.8 $\pm$ 4.5  & \textbf{85.1 $\pm$ 4.2} & \textbf{85.6 $\pm$ 4.6} \\ 
      {JPN}    & 93.4 $\pm$ 2.4  & 93.2 $\pm$ 3.2 & \textbf{96.5 $\pm$ 2.2*}\\ 
      {LET}     & 86.6  $\pm$ 1.5 & \textbf{95.2 $\pm$ 0.7} &  \textbf{95.1 $\pm$ 0.7} \\
      {MNT}  & 91.5 $\pm$ 0.2 & 84.5 $\pm$ 0.1 & \textbf{93.9 $\pm$ 0}\\
      
       \hline
             &  ULFMM~\cite{Figueiredo2002}  & MoG-D & MoG-F \\
        
         \hline
            {ORL}   & 80.0 $\pm$ 6.5  & 89 $\pm$ 2.4  & 90 $\pm$ 0  \\
            {LVQ}   & 75.4 $\pm$ 4.5 &  88.1 $\pm$ 2.6 & \textbf{92.1 $\pm$ 1.8} \\
            {OPT}  & 49.5 $\pm$ 10.2  &  84.2 $\pm$ 3.1 & \textbf{94.9 $\pm$ 1.7} \\
            {PEN} & 89.9 $\pm$ 2.0 & 84.5 $\pm$ 2.0 & 97.4 $\pm$ 0.6\\
            {VIS}   & 20.6 $\pm$ 3.7 & 68.6 $\pm$ 3.9 & 44.7 $\pm$ 12.8  \\
            {WAV}   & 72.1 $\pm$ 7.5  & 80.9 $\pm$ 18.2 & \textbf{84.8 $\pm$ 4.6}  \\ 
            {JPN}   & 82.4 $\pm$ 2.1  & 82.2 $\pm$ 4.9 & 92.3 $\pm$ 2.3\\ 
            {LET}   & 56.9 $\pm$ 2.8 & 64.2 $\pm$ 1.2 &  88.6 $\pm$ 0.9 \\
            {MNT}   & 64.7 $\pm$ 2.0 & 78.2 $\pm$ 0 & \textbf{ 93.7 $\pm$ 0}\\
   \hline
   \end{tabular}%
 \label{tab:allresults}%
\end{table}%

The best results for a dataset are shown in \textbf{bold}. We compared the algorithms with a non-parametric sign test. 
For each dataset, we conducted a one tail paired-sample t-test with a significance level of 0.05 (0.01 upon of rejection of null hypothesis). Results indicate that ULFMM ranks last in all cases: it is consistently inferior even against the MoG-F baseline. This is because of the fact that after randomized initialization of clusters, ULFMM algorithm annihilates all illegitimate components, skipping intermediate (possibly better than initial) models. On seven datasets AMoFA attains/shares the first rank, and on the remaining two it ranks the second. Note that though on the overall AMoFA and VBMoFA have similar number of wins against each other, on high dimensional datasets, namely on MNIST, VISTEX, Japanese Phoneme and ORL, AMoFA significantly outperforms VBMoFA. 

\begin{table}[htbp]
  \centering
      \caption[Row Wins/Ties/Loses]{Row Wins/Ties/Loses against Column with 0.05 Significance. }
    \begin{tabular}{|r|c|c|c|c|c|}
	\hline
           & AMoFA  & VBMoFA & ULFMM & MoG-D & MoG-F \\
    \hline
    IMoFA       & 1/2/6  & 2/4/3& 9/0/0 & 7/2/0 & 2/3/4 \\ \hline
    AMoFA       &    *   & 4/3/2& 9/0/0 & 7/2/0 & 5/3/1 \\ \hline
    VBMoFA      &        & *    & 9/0/0 & 7/2/0 & 3/5/1 \\ \hline
    ULFMM       &        &      &   *   & 0/3/6 & 0/0/9 \\ \hline
    MoG-D		&        &      &       &  *    & 1/2/6 \\
    \hline
    \end{tabular}%
  \label{tab:pairwisewls}%
\end{table}%

The results of pairwise tests at 0.05 significance level are shown in Table~\ref{tab:pairwisewls}. We see that the adaptive MoFA algorithms dramatically outperform GMM based ULFMM algorithm. 
MoFA is capable of exploring a wider range of models between diagonal and full covariance with reduced parameterization. Among the three MoFA based algorithms, no significant difference ($\alpha=0.05$) was found on Pendigits dataset. AMoFA outperforms the best results reported so far with the VISTEX dataset. The best test set accuracy reported in~\cite{Salah04imofa} is 73.8 $\pm$ 1.1 using GMMs. 
We attain 77.2 $\pm$ 4.6 with AMoFA.  
\pagebreak
\section{Conclusions and Outlook}
\label{lbl:conclusions}
In this study, we propose a novel and adaptive model selection approach for Mixtures of Factor Analyzers. Our algorithm first adds factors and components to the mixture, and then prunes excessive parameters, thus obtaining a parsimonious model in a very time and space efficient way. 
Our contributions include a generalization of the adopted MML criterion to reflect local parameter costs, as well as local component annihilation thresholds. 

We carry out experiments on many real datasets, and the results indicate the superiority of the proposed method in class-conditional modeling. We contrast our approach with the Incremental MoFA approach~\cite{Salah04imofa}, Variational Bayesian MoFA~\cite{Ghahramani1999}, as well as the popular ULFMM algorithm~\cite{Figueiredo2002}. 
In high dimensions, MoFA based automatic modeling provides significantly better classification results than GMM based ULFMM modeling, as it is capable of modeling a much wider range of models with compact parametrization. It also makes use of the latent dimensionality of the local manifold, thus enables obtaining an adaptive cost for the description length. AMoFA algorithm is observed to offer the best performance on higher dimensional datasets. 

The proposed algorithm does not necessitate a validation set to control model complexity. Thanks to the optimized MML criterion and the fast component selection measures for incremental adaptation, the algorithm is not only robust, but also efficient. It does not have any requirement for parameter tuning.
Using a recursive version of ULFMM~\cite{Zivkovic04RULFMM}, it is also possible to extend the proposed method for online learning.  A MATLAB tool for AMoFA is available from~\url{http://www.cmpe.boun.edu.tr/~kaya/amofa.zip}. 


\vfill \pagebreak

\appendix
\section{EM Algorithm for Mixture of Factor Analyzers with MML Criterion}
\label{appsec:app_em_mofa}
In this section, we give the MoFA EM algorithm optimizing the generalized MML criterion given in eq~\ref{eq:FJMMLAdoptedAmoFAFinal}. This criterion is used for automatic annihilation of components at the M step. We provide the formulation of MML based EM algorithm, which is closely related to regular EM for MoFA model~\cite{Ghahramani97EM4MFA}: 
\begin{eqnarray}
\label{eq:MoFA_EMez}
E[\bz|\cG_k,\bx^t] & = &
h_{ik}\Omega_k\left(\bx^t - \bmu_k \right)
\\
\label{eq:MoFA_EMezz}
{E\left[\bz\bz'|\cG_k,\bx^t\right]} & = & 
\begin{split}
&h_{ik} (I- \Omega_k\bLambda_k
+\Omega_k(\bx^t-\bmu_k)(\bx^t-\bmu_k)'\Omega{'})
\end{split}
\\
\label{eq:MoFAEMLambdaMu}
\tilde{\bLambda}_k^{\text{new}} & = &  
\begin{split}
&\left( \sum_i h_{ik} \bx^t E\left[\tilde{\bz}|\bx^t,\cG_k\right]' \right)
\left( \sum_j h_{jk} E[\bz\bz'|x_j,\cG_k]  \right)^{-1} 
\end{split}
\\
\label{eq:MoFAEMPsi}
\bPsi_k^{new} & = & 
\begin{split}
&\frac{1}{N \pi_k} diag \{ \sum_{i}h_{ik}(\bx^t -
 \tilde{\bLambda}_k^{new} E[\tilde{\bz}|\bx^t,\cG_k])\bx^t{'} \}
\end{split}
\\
\label{eq:MoFAEMPi}
\pi_k^{\text{new}} & = & \frac{1}{N} \sum_{i=1}^{N} h_{ik}
\end{eqnarray}
where to keep the notation uncluttered, $\tilde{\bz}$ is defined as $ \begin{bmatrix} \bz & 1 \end{bmatrix}'$. Similarly,   $\tilde{\bLambda}_k =  \begin{bmatrix} \bLambda_k & \bmu_k \end{bmatrix}$, $\Omega_k \equiv \bLambda_k (\bPsi_k + \bLambda_k\bLambda_k{'})^{-1}$, and
\begin{equation}
\label{eq:MoFA_EMeh}
 h_{ik} = {E\left[\cG_k|\bx^t\right]} \propto p(\bx^t,\cG_k)= \pi_k {\cal N}\left(\bmu_k, \bLambda_k\bLambda_k{'}+ \bPsi_k\right). 
\end{equation}
The above EM formulation aims to optimize the MoFA log likelihood, which is the logarithm of the linear combination of component likelihoods:
\begin{equation}
\begin{split}
 p\left( {\cal X}|\bz,\cG \right) & =log \prod_{i=1}^N \sum_{k=1}^{K} \pi_k {\cal N}\left(\bx^t;\bmu_k, \bLambda_k\bLambda_k{'}+ \bPsi_k\right)
\end{split}
\label{eq:MFAloglik}
\end{equation}
\begin{figure}[htbp]
\caption[EM Algorithm for MoFA]{\label{alg:mofa_em_mml}EM Algorithm for MoFA  with MML Criterion}
\begin{center}
\framebox[5.0in]{\begin{minipage}[t]{4.9in}
\begin{algorithmic}
\STATE \textbf{Require: ${\cal X}$} data, and initialized MoFA parameter set $\btheta=\{\bmu,\bLambda,\bPsi,\pi\}$ 
\STATE \textbf{REPEAT}
\STATE   E Step: compute expectations $h_{ik}, \text{E}[\bz|\cG_k,\bx^t], {E\left[\bz\bz'|\cG_k,\bx^t\right]}$ using eq.\ (\ref{eq:MoFA_EMeh}), (\ref{eq:MoFA_EMez}) and (\ref{eq:MoFA_EMezz}), respectively
\STATE   M step: compute model parameters using equations (\ref{eq:MoFAEMLambdaMu})-(\ref{eq:MoFAEMPi})
\STATE   Compute $T_k = C_k/2$ using eq.\ (\ref{eq:compannihthresh})
\STATE   \textbf{while} any component needs annihilation
\STATE    Annihilate \textbf{the weakest} component $k$ having  $N_k < T_k$
\STATE    Update $\pi_k = \pi_k / \sum_{l=1}^{K_{nz}^{new}} \pi_l , 1 \leq k \leq K_{nz}^{new} $
\STATE   \textbf{end}
\STATE   Compute ${\log p({\cal X}|\btheta})$ using eq.\ (\ref{eq:MFAloglik})
\STATE   Compute message length ${{\cal L}(\btheta,{\cal X})}$ using eq.\ (\ref{eq:FJMMLAdoptedAmoFAFinal})
\STATE   \textbf{UNTIL} ${{\cal L}(\btheta,{\cal X})}$ converges with $\epsilon$ tolerance

\end{algorithmic}
\end{minipage}}
\end{center}

\end{figure}
The EM Algorithm for MoFA using MML criterion is given in figure~
\ref{alg:mofa_em_mml}. 

\bibliographystyle{elsarticle-num}
\bibliography{references}
\end{document}